# Toward Agentic AI-Ready Workflows in Built Environment Organizations: Knowledge Structures and Interoperability Challenges

**A. B. Ersoz**
***Institute for Environmental Design and Engineering, The Bartlett, Faculty of the Built Environment, University College London, London, UK***
*a.ersoz@ucl.ac.uk*

**J. McHenry**
***GIA Surveyors, London, UK, School of Natural and Built Environment, Queen's University Belfast, UK***
*justin.mchenry@gia.uk.com*

**Y. Kovalchuk**
***Centre for Advanced Research Computing, University College London, London, UK***
*y.kovalchuk@ucl.ac.uk*

**E. M. Barrett, M. S. Khanie**
***Institute for Environmental Design and Engineering, The Bartlett, Faculty of the Built Environment, University College London, London, UK***
*edward.barrett@ucl.ac.uk, m.khanie@ucl.ac.uk*

## Abstract

*The rise of agentic artificial intelligence (AI) in the built environment is shifting attention from isolated AI tools toward autonomous systems that reason, plan, and act across project and organizational workflows. The effectiveness of such systems depends not only on large language model capabilities, but also on the readiness of heterogeneous data environments across the Architecture, Engineering, and Construction (AEC) sector. These environments combine structured sources, such as CAD and BIM models, schedules, cost databases, and asset registers, with unstructured ones, including technical reports, design narratives, planning and policy documents, contracts, meeting minutes, and risk registers. This paper reviews how built environment data can be organized to support agentic AI-ready workflows, focusing on knowledge structures, interoperability, and access and integration mechanisms such as retrieval-augmented generation (RAG) and the Model Context Protocol (MCP). Drawing on existing literature, it examines approaches proposed to organize structured and unstructured data and considers how such data can be contextualized and made accessible to AI agents in a reliable manner. The paper synthesizes current perspectives, highlights challenges around fragmentation and interoperability, and identifies open questions for future research on AI-compatible information management in the built environment.*

# Introduction

Artificial intelligence (AI) in the built environment is undergoing a transformative shift. For most of the past decade, AI entered Architecture, Engineering, and Construction (AEC) practice as a collection of isolated statistics-based tools. For example: a classifier for identifying defect images, a regressor for predicting cost overruns, or a chatbot for semantic document search. What is now appearing is a different class of system: agentic AI, in which large language models (LLMs) serve as the reasoning core of agentic tools that perceive their environment, maintain memory, decompose goals into plans, invoke external tools, and act with limited human supervision (L. Wang et al., 2024). Increasingly, such agents are organized into multi-agent configurations whereby specialized roles coordinate, negotiate, and check one another's work; however, the reliability of collective decision-making in these systems remains an open problem (Sun et al., 2025). Within the built environment, this trajectory has been consolidated under the concept of agentic built environments, where autonomous agents operate continuously across design, construction, and operations workflows rather than answering isolated queries (Lee et al., 2025). The shift is not only technical, but also relates to the platform-based delivery of AI capability in a way that is beginning to reshape the business models of construction organizations themselves (Nyqvist et al., 2025).

The agentic AI transition depends on a precondition that is often overlooked: data readiness. The point is not that data matters more than model capability; model-side limits remain real, and the two interact, since more capable models can work with less-prepared sources. Yet in practice, implemented systems keep failing at the data boundary (K. Liu et al., 2025; Zhu & Liu, 2025). General-purpose models hallucinate domain content and cannot act on project data unless it is explicitly supplied (Prieto et al., 2023; Saka et al., 2024), and recent reviews point to data access, validation, and integration as the primary barriers to deployment (Gao, Yiu, et al., 2026; Park et al., 2026). Which of these constraints weighs more cannot yet be measured, because shared benchmarks do not exist. Moreover, there is no common legislative or policy environment governing and guiding the transition to AI integration within the various distinct domains of built environment professions, e.g. planning, architecture, engineering, and construction (Agapiou, 2024). The literature shows that every demonstrated systems had to prepare its own data before its agents could act. This makes readiness a management concern as much as a technical one. It is specified, procured, owned, and paid for through the sector's information-management practices, and it therefore belongs on the agenda of project and construction management.

The data environments that agents must act on are highly heterogeneous, including both structured sources (CAD and BIM models, schedules, cost databases, asset registers) governed by their own schemas, and unstructured sources (reports, planning documents, contracts, meeting minutes, risk registers), where most organizational knowledge actually belongs (Hassan et al., 2021), and thus are essential to the unique operation of each business. Generally, the semantic links between the two types of data sources are mostly left implicit (Sobhkhiz & El-Diraby, 2023). In short, built environment data is organized for human consumption within disciplinary boundaries, not for machines to read across them (Chen et al., 2025; H. Wang et al., 2024; Yang et al., 2024). The same divide runs through the wider information sources of built environment organizations. The enterprise records of clients, operators, and the professional consultancies that serve projects (fee proposals, client and contact records, correspondence, survey and analysis outputs) sit alongside project data, yet are typically no more machine-traversable than unstructured data due to the organic way in which organisations' systems, processes and procedures evolve over time.

This paper therefore reviews how built environment data can be organized to support agentic AI-ready workflows. It considers three research questions:

1. How are structured and unstructured built environment data currently organized for machine readability?
2. Through which mechanisms is such data contextualized and made accessible to AI agents?
3. What does block reliable agentic access to built environment data, and what should future research prioritize?

This is a narrative review conducted through a structured search and analysis process. The review is based on a corpus of 69 studies published between 2012 and 2026. Candidate papers were first identified through iterative keyword searching of Google Scholar, Scopus and Web of Science, using combinations of terms covering agentic AI, LLMs, BIM and construction, knowledge graphs, retrieval-augmented generation, and the Model Context Protocol. The same combinations of terms were applied across all the databases, adapted to the syntax of each platform. The set was consequently extended by following the references of recent reviews and later publications citing them. Studies were retained if they concerned the organization, representation, or serving of built environment data to LLM-based systems, or if they provided the domain-agnostic agent and protocol foundations needed to frame that discussion. Each study was classified into thematic clusters and examined through the three-layer framework shown in Figure 1. We derived the three layers from the thematic clustering of the literature, based on our own interpretation and synthesis of reviewed studies. Across the reviewed papers, three concerns appeared as separate: the condition of the data, the structures used to attach meaning to it, and the mechanisms through which agents reach it. The data layer asks what sources exist and in what condition; the knowledge structure layer asks how meaning is attached to those sources so that agents can traverse them; and the access and integration layer asks through which mechanisms an agent reaches them. The literature on agentic applications is mapped against all three layers throughout the paper. Section 2 presents the findings layer by layer, Section 3 discusses challenges and implications, and Section 4 concludes the paper.

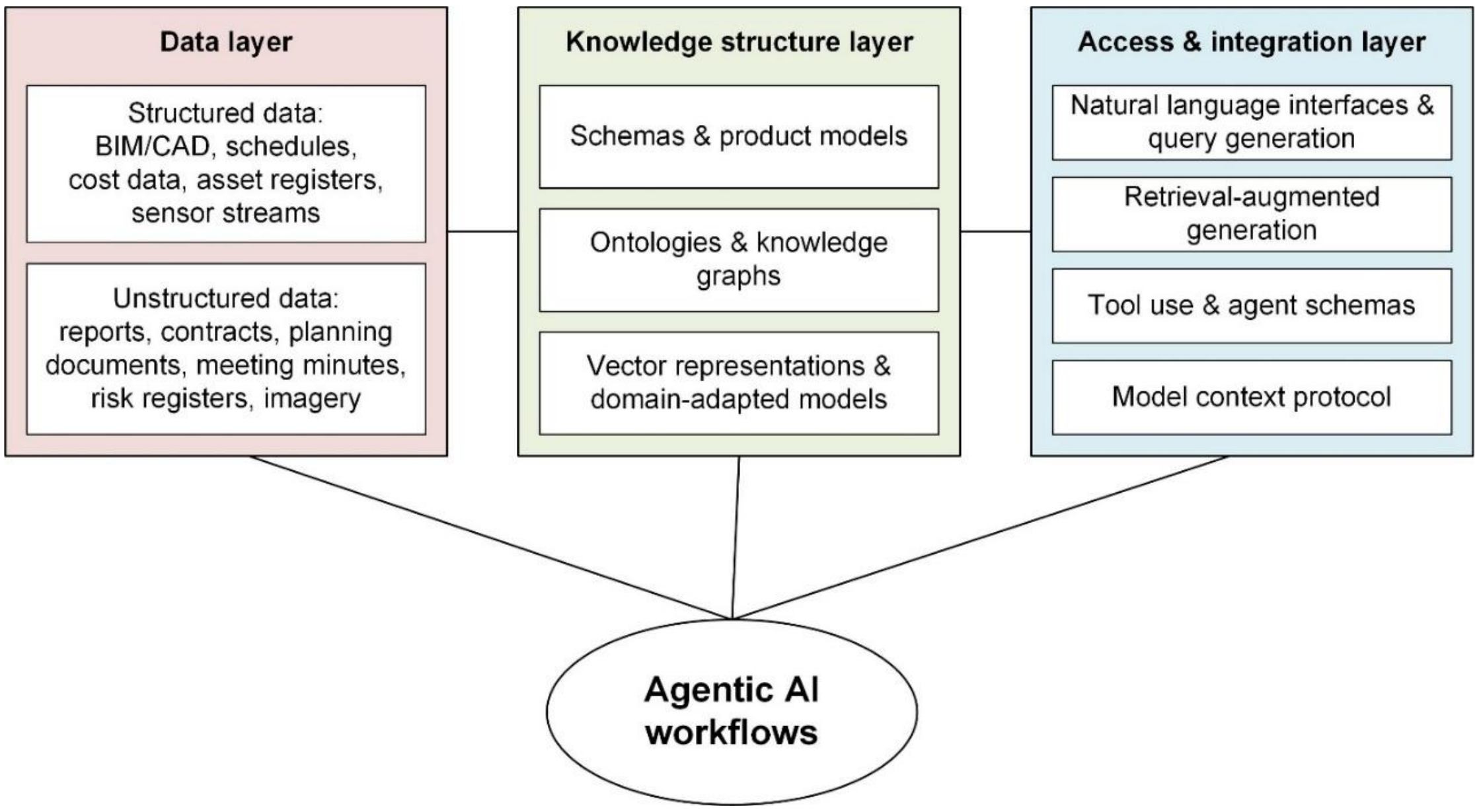

Figure 1. Three-layer framework adopted in this review for analysing literature

# Findings

The findings are structured along the three-layer framework. Together, the findings across the layers outline what it takes for an agent to act on built environment data reliably.

## The built environment data landscape

The first layer of the framework (Figure 1) addresses the data itself: what sources exist and in what condition. Structured data in the AEC industry (BIM and CAD models, schedules, cost databases, asset registers, sensor streams, enterprise systems such as ERP and client records, and public registers such as planning and land administration data) is defined by schemas. An IFC model, for example, provides typed entities and relationships that software can parse deterministically, although in practice, exchange reliability depends on model view definitions and exporter quality. Schema compliance, however, does not guarantee that meaning is fully captured. Multi-constraint queries require a supplementary ontology layer beyond what the model encodes (Gao, Hu, et al., 2026; Yin, Tang, Webster, Xu, et al., 2023).Models evolve through asynchronous editing by multiple parties, so the "current state" an agent queries may be undefined without explicit version control (Esser et al., 2022). In contrast, unstructured data (technical reports, planning and policy documents, contracts, meeting minutes, risk registers, site imagery) holds most of the sector's decision-relevant knowledge and nearly all of its legal commitments; however, its meaning sits in free text, tables, and images that must be interpreted before a machine can use it (Tixier et al., 2016; Wong et al., 2024). The two types of data are maintained by different roles in different systems, and the links between them are rarely made explicit. Integrating them requires either laborious manual mapping or machine-learning approaches that bypass the schema entirely (Sobhkhiz & El-Diraby, 2023). Enrichment and knowledge-graph research exist precisely to create these missing links (Forth & Borrmann, 2024).

Based on patterns that recur across the corpus, we present the following six conditions that indicate how ready an information source is for reliable agentic use:

- machine-readability - can a specific item be retrieved programmatically?
- semantic explicitness - is meaning encoded or merely implied?
- provenance and versioning - is it clear where the data came from and which state is current?
- granularity - can the agent retrieve information at the level it reasons at?
- update cadence - does the source reflect present reality? and
- access control - can permissions be enforced at the point of agentic access?

Table 1 applies these six conditions to the main source types. Cell ratings (low, moderate, and high) are based on the authors' qualitative judgments of common current practice, compiled from the corpus rather than extracted from specific sources. Rating anchors:

- High: the property is supplied by the source's native schema or governance regime;
- Moderate: supplied inconsistently or only for part of the content;
- Low: largely absent in typical practice.

Table 1. Agent-readiness of built environment data sources in current practice

| Source type | Machine-readability | Semantically explicit | Versioned/ provenanced | Access control at agent level | Granularity | Update cadence |
|---|---|---|---|---|---|---|
| BIM / IFC models | High | Moderate | Moderate | Low | Element-level | Irregular |
| Schedules /cost databases | High | Moderate | Moderate | Low | Activity or line-item | Irregular |
| Asset registers | High | Moderate | Low | Low | Asset | Slow |
| Sensor/ BMS streams | High | Low | Moderate | Low | Point | Real-time |
| Contracts /legal documents | Low | Low | High | Moderate | Clause | Static |
| Reports, narrative, minutes | Low | Low | Low | Moderate | Clause | Varies |
| Planning / policy documents | Low | Low | Moderate | Moderate | Clause | Slow |
| Risk registers | Moderate | Moderate | Low | Low | Entry | Irregular |
| Site imagery | Moderate | Low | Low | Low | Image | Frequent |

**Knowledge structures for agent-ready data**

The second layer of the proposed framework (Figure 1) addresses knowledge structures that are the ways in which meaning is attached to data so that agents can act on it. For structured data, the sector's standard answer is a common data schema. Schemas of this kind give agents a formal target against which their outputs can be validated; however, schemas alone do not settle meaning. For example, systems built on IFC still need schema-guided reasoning layers and curated tool interfaces before agents can operate on them reliably (Yin, Tang, Webster, Li, et al., 2023). Ontologies and knowledge graphs go a step further by making the meaning of data explicit and connected. In the built environment, they have been used to construct a domain-wide knowledge graph from research literature (Yang et al., 2024), to share safety knowledge across organizations (Fang et al., 2020; Pedro et al., 2022), and to serve as the world model through which agents interpret digital twins (Yoon et al., 2025). What a graph gives an agent, compared with a document store, is relational reasoning: the ability to follow links, for example, from an observed fault to its cause and then to the responsible system, which agentic safety systems now rely on (A. Li et al., 2026; Speiser et al., 2026).

Between authored structure and free text sits a third option: enrichment research, which provides missing semantics after the fact. Machine learning, rule-based inference, and textual similarity can recover information that authoring practice failed to record (Forth & Borrmann, 2024), and unstructured content can be attached to model elements without a shared schema (Sobhkhiz & El-Diraby, 2023). Embedding-based indexes and domain-adapted language models advance this direction further, trading explicitness for coverage. Anything can be embedded, but nothing is verifiable by inspection. Pre-training on construction corpora measurably improves domain performance (Zhong & Goodfellow, 2024), but knowledge stored in parameters becomes outdated as soon as training ends and cannot point back to its sources. Hybrid vector-graph retrieval, for example in BuildingGPT (M. Li & Wang, 2026), points to the likely way forward.

These three structural families (schemas, ontologies and graphs, and vector representations) serve different agent operations: schemas support precise, validatable retrieval; graphs support relational reasoning; vectors support semantic recall over unrestricted content (Lee et al., 2024). Across the corpus, the observed norm is already hybrid, so the design question is how to compose these structures, not which one to choose. For a project organization, the practical implication is clear: each of these structures is an information asset that someone must create and maintain. Ontologies and enrichment pipelines do not emerge on their own as a by-product of design and construction work; in the demonstrations reviewed, the research teams built them themselves.Efforts to make existing ontologies discoverable and reusable, such as the BE-OLS (Built Environment Ontology Lookup Service), are a step toward reducing this duplication (Akbarieh et al., 2025). However, reuse still depends on each project selecting, adapting, and maintaining a structure of its own.

### Access and integration mechanisms

The earliest access mechanism in the corpus is the natural-language interface to structured project data. Over fifteen years, these interfaces have progressed from template-based question answering through learned semantic parsing to LLM-native pipelines (Lin et al., 2016; D. Liu et al., 2025; Yin, Tang, Webster, Li, et al., 2023). What changed at each step was the interpreter. What never changed was the need for a formal target into which language is translated, whether a query language or executable code. Recent studies treat this as a clear design principle: constraining the model to a purpose-built domain-specific language or a curated code library keeps its output within a space that can be verified (Guo et al., 2025; Koh et al., 2026). Query generation is therefore best understood as a grounding strategy. While the model proposes, a formal executor decides what actually runs.

Retrieval-augmented generation (RAG) is another retrieval mechanism. In a RAG pipeline, documents relevant to the query are retrieved first and supplied to the model as context, so the output is grounded in that evidence rather than in what the model memorized during training. In construction safety management, comparative evaluation finds RAG preferable to model fine-tuning on accuracy, currency, and auditability grounds, and applied systems show that retrieval over curated regulatory and incident corpora improves the specificity of generated guidance (Baek et al., 2025; Speiser et al., 2025). However, the quality of RAG is constrained by how the corpus is organized: basic vector retrieval struggles with relational questions, which has motivated vector-graph hybrids (M. Li & Wang, 2026), and retrieval cannot surface links that were never made explicit. That limitation leads back to the knowledge structures discussed in the previous section.

Not all built environment data can be retrieved as text; much of it must be operated on through software. Tool use gives agents this capability through functions with typed signatures, and the domain evidence shows that designing these tool interfaces is itself a data-organization problem. The granularity of the tools must match the steps of the workflow the agent reasons about (Xu et al., 2025). Designing the tool layer also means deciding in advance which operations on project data are safe, meaningful, and composable. That decision is a governance act as much as an engineering one. Whoever writes the tool schema decides what an agent may do to the project record, and no current role description assigns that authority.

The Model context protocol (MCP) standardizes how agents discover and invoke external context and tools. Servers expose resources and typed operations, and AI agents can use them without bespoke integration (Ayyagari, 2025). For a fragmented sector, the importance is structural. The MCP proposes to do for agent-system integration what IFC proposed for model exchange, but at the level of runtime access rather than file exchange. First AEC applications are appearing, from MCP-wrapped IFC design operations (Nithyanantham et al., 2025) to MCP-orchestrated building energy modelling workflows (H. Li et al., 2026), although the evidence base for a protocol only two years old is still limited.

Table 2 compares the access and integration mechanism families and reinforces a layered rather than competitive reading: the MCP and tool use standardizes transport and discovery; query generation and RAG are retrieval strategies that can ride on either; and all three are fundamentally limited by the knowledge structures described in the previous section (M. Li & others, 2026). While empirical studies are oriented toward BIM-centric, design- and construction-phase pipelines, the mechanisms themselves are representation-agnostic. Query generation, RAG, tool schemas, and MCP apply unchanged to the enterprise, geospatial, and regulatory data that dominate the information environments of consultancies and asset operators.

Table 2. Comparison of access and integration mechanisms for agentic use of built environment data.

| **Dimension** | **Query generation** | **Retrieval-augmented generation** | **MCP tool use** |
|---|---|---|---|
| Primary object | Structured data stores | Document corpora | Exposed resources and tools |
| Verifiability of access | High | Moderate | High |
| Write capability | Moderate | None | High |
| Grounding quality | High | Moderate | Moderate |
| Independence from source system | Low | High | High |
| Governance / security maturity | High | Moderate | Moderate |

## Discussion

The findings demonstrate that technical building blocks for agentic workflows already exist. What stands between these building blocks and reliable practice is less clear, and that gap is the subject of this section. Three issues arise from the literature review and are discussed below in turn, together with our interpretation of the findings and the implications we draw from them.

### Fragmentation, interoperability, and the readiness

Fragmented and closed solutions is the recurrent theme from the findings: each reviewed system solved interoperability locally, for its own data and its own use case, and nothing carried the solution further. Looking closely, the interoperability challenge is in fact three distinct problems. Format-level interoperability (can systems parse each other's files?) appears comparatively well addressed across the reviewed studies, through schemas and exchange standards. Semantic interoperability (do systems agree on what the data means?) is addressed only within isolated applications or domains in the reviewed studies: linked-data safety systems, domain-scale knowledge graphs, shared dictionaries, and enrichment pipelines, each establish meaning locally, but nothing carries it across the wider project ecosystem. Organizational interoperability (will firms open their data to one another's agents at all?) is barely addressed in the corpus. Whether firms do open their data may depend as much on the business models of platform providers as on protocols. The MCP addresses the format level and, in part, the semantic layer of runtime access. It does not decide which organization's data an agent should treat as current and correct, a question that version-control research shows to be challenging even within a single model. The benefits identified in the existing literature remain largely confined within individual organizations, primarily through improvements in information retrieval and automated processes. On this basis, we suggest that greater value could emerge when agents are able to operate across organizational and professional boundaries, which would also require data to flow across those boundaries. In our view, publicly available datasets may provide a practical starting point for this form of interoperability. Planning systems and land administration records, for example, are already used across multiple professions in the sector. Making these datasets machine-readable and accessible to agents could provide a shared foundation that would be difficult for any single firm to establish independently.

The organizational problem also matters because readiness has so far been created one project at a time. In the demonstrations reviewed, knowledge structures and tool layers were built for each use case by the system authors themselves. Whether readiness can instead be supplied at sector scale is an economic question, and the sector has faced it before. IFC's promise of semantic interoperability underdelivered for decades largely for economic rather than technical reasons, and research on industry-wide standards explains why: shared semantics benefit everyone, but their costs fall on individual firms, and a fragmented, project-based industry lacks the long-term relationships that would justify investing in shared infrastructure (Papadonikolaki, 2018). The same conditions hold for the MCP in AEC, so the MCP-IFC analogy should be read as a warning rather than a forecast. Without procurement mandates or platform incentives that change these economics, a technically robust protocol can stall exactly where its predecessor did. It is not even certain that investing in readiness pays off. Earlier attempts to encode engineering knowledge in machine-usable form, most notably knowledge-based engineering, failed on acquisition cost, drift from practice, and brittleness (Pokojski et al., 2022; Rocca, 2012), and some vagueness in project information is left there on purpose, serving negotiation and liability management. There is also a rival technical path. Instead of making data agent-ready, make agents tolerant of unready data, using machine learning to prepare it on demand. What separates the present moment from those earlier attempts is that the interpreting model is now cheap and general, and part of the encoding work can be automated. Whether that changes the economics of adoption is an unresolved empirical question, and any sector-scale readiness agenda depends on the answer.

### Reliability and evaluation

Hallucination in construction settings is not a theoretical risk. General-purpose models generate domain content when detached from project data. The mechanisms discussed in this paper are, collectively, countermeasures against this risk. RAG supplies evidence, and query generation confines outputs to verifiable spaces. Yet, these mechanisms relocate the reliability question rather than eliminate it, and multi-agent configurations add a further layer, because consensus among agents can amplify rather than correct individual errors. Earlier decision-support research offers a practical standard. Multi-criteria frameworks treated interactive explanation as a design requirement rather than an afterthought (Buron Brarda et al., 2023), and agentic systems acting on built environment data should be held to at least that standard. Evaluation practice makes matters worse. Ad-hoc case demonstrations dominate over comparable evaluation, and construction suffers shared benchmarks that have driven cumulative progress in the mainstream natural language processing (NLP) field. Until such benchmarks exist, claims about agentic readiness, including the qualitative characterizations offered in this review, are difficult to be independently tested.

### Security, governance, and human oversight

Agentic access concentrates risk in a way document access never did. Security research on MCP has already identified several specific threats (Hou et al., 2025). These include tool poisoning, where misleading tool descriptions influence an agent's behavior. Another risk is prompt injection through server-provided content. Agents may also gain permissions beyond those originally intended. In an industry where data is commercially sensitive and contractually confidential, an agent holding standing credentials across project systems is therefore a qualitatively new liability, and governance must operate at the tool layer, deciding which operations exist at all, rather than only at the identity layer. Empirical studies converge on a complementary safeguard, keeping humans structurally inside the loop. However, security is only half of governance. The other half concerns people and liability, and the reviewed literature is largely silent on it. Several reviewed systems continuously capture worker speech and site imagery (Pu et al., 2024), a regulated activity that raises consent and workplace-surveillance questions, especially on sites with multiple employers; the lawful basis for such capture and reuse may deserve a place among the readiness conditions in Table 1. Write capability raises a related question: none of the reviewed studies address who bears professional and legal liability when an agent modifies a model, schedule, or register. Finally, the workflows these systems automate are currently performed by document controllers, planners, and information managers, and this workforce transition is an adoption factor that the technical literature does not yet acknowledge. Moreover, the application of these systems within distinct strands of the built and natural environment professions and the different specialist areas and roles performed by humans remains an underexplored area.

## Conclusion

This paper examined how built environment data can be organized to support agentic AI-ready workflows across the data landscape, knowledge structures that make data meaningful, and access mechanisms through which agents reach it. The three research questions can now be answered. The first question asked how structured and unstructured built environment data are currently organized for machine readability. The reviewed studies demonstrate data organized for human use: structured data are schema-defined but semantically incomplete, unstructured data hold most decision-relevant knowledge in that require interpretation and contextualization before they can be used reliably by agents. The links between the two sources are rarely explicit. Across the reviewed demonstrations, the condition and organisation of the underlying data emerged as a recurrent constraint on agentic capability alongside limitations in model competence. Readiness was manufactured bespoke rather than inherited from the information environment. The second question asked through which mechanisms such data is contextualized and made accessible to AI agents. Knowledge structures and access mechanisms are complementary layers rather than alternatives: schemas, graphs, and vectors serve different agent operations, and RAG, query generation, tool schemas, and MCP complement each other rather than compete, with the MCP being the least mature of these mechanisms. The third question asked what blocks reliable agentic access and what future research should prioritize. The fragmentation that blocks agentic workflows is not only technical but also organizational. The review did not explicitly identify an established role, process, or contractual mechanism responsible for creating and maintaining semantic links between the structured and unstructured data. There are two further barriers. Without shared benchmarks, claims about agentic readiness cannot be independently tested. Governance responses to agentic access risks remain underdeveloped, particularly around permissioning, professional and legal liability, human oversight, and responsibility for agent-initiated changes. Making the built environment agent-ready, for delivery teams and for the professional organizations that serve them, is therefore an information-management agenda before it is an AI agenda. Three directions for future research follow from this review. First, the readiness conditions proposed in this paper should be developed into measurable criteria and shared benchmarks, so that agentic systems can be compared rather than merely demonstrated. Second, deployment studies on real projects are necessary to establish what readiness costs and what it returns. Third, the responsibility for readiness remains unassigned, and designing the necessary roles and contractual instruments is a task for project and construction management research, alongside understanding whether practitioners across the sector are ready to adopt these tools in their daily work.

## Acknowledgements

This research was funded by UK Research and Innovation (UKRI) through Innovate UK, as part of a Knowledge Transfer Partnership (KTP no. 10148562) between University College London and GIA Surveyors.